\documentclass[10pt,twocolumn,letterpaper]{article}

\usepackage{cvpr}
\usepackage{times}
\usepackage{epsfig}
\usepackage{graphicx}
\usepackage{amsmath}
\usepackage{amssymb}
\usepackage{subfig}
\usepackage{multirow}
\usepackage{algorithm}
\usepackage{algorithmic}
\usepackage[titletoc]{appendix}
\usepackage[breaklinks=true,bookmarks=false]{hyperref}

\cvprfinalcopy % *** Uncomment this line for the final submission

\def\cvprPaperID{****} % *** Enter the CVPR Paper ID here

\begin{document}

%%%%%%%%% TITLE
\title{BEVPoolv2: A Cutting-edge Implementation of BEVDet Toward Deployment}

\author{Junjie Huang\thanks{Corresponding author.} \quad Guan Huang \\
       PhiGent Robotics\\
       {\tt\small {junjie.huang,guan.huang}@phigent.ai}
%\and
%Zheng Zhu\\
%Beijing University of Posts and Telecommunications, Beijing\\
%Beijing, China\\
%{\tt\small zhengzhu@ieee.org}
}

\maketitle
%\thispagestyle{empty}

%%%%%%%%% ABSTRACT
\begin{abstract}
We release a new codebase version of the BEVDet, dubbed branch dev2.0. With dev2.0, we propose BEVPoolv2 upgrade the view transformation process from the perspective of engineering optimization, making it free from a huge burden in both calculation and storage aspects. It achieves this by omitting the calculation and preprocessing of the large frustum feature. As a result, it can be processed within \textbf{0.82 ms} even with a large input resolution of $640\times1600$, which is \textbf{15.1 times} the previous fastest implementation. Besides, it is also less cache consumptive when compared with the previous implementation, naturally as it no longer needs to store the large frustum feature. Last but not least, this also makes the deployment to the other backend handy. We offer an example of deployment to the TensorRT backend in branch dev2.0 and show how fast the BEVDet paradigm can be processed on it. Other than BEVPoolv2, we also select and integrate some substantial progress that was proposed in the past year. As an example configuration, BEVDet4D-R50-Depth-CBGS scores 52.3 NDS on the NuScenes validation set and can be processed at a speed of 16.4 FPS with the PyTorch backend. The code has been released to facilitate the study on \url{https://github.com/HuangJunJie2017/BEVDet/tree/dev2.0}.
\end{abstract}

\section{Introduction}
The task of multi-camera 3D object detection is fundamental in autonomous driving, which draws great attention in both the research and the industry community recently. Since the scalable paradigm BEVDet \cite{BEVDet} was first introduced to this problem, we have witnessed a mass of progress \cite{BEVDet4D, BEVDepth, BEVStereo, BEVFusionMIT, BEVFusion, UVTR, STS, SOLOFusion, BEVerse} in the past year. With this technical report and the released code, we offer a new codebase version of the BEVDet, dubbed dev2.0. BEVDet-dev2.0 is deployment oriented with an outstanding trade-off between inference speed and accuracy. It inherits the advantages of the BEVDet \cite{BEVDet} while is further developed from several aspects like engineering optimization of some modules and feature integration of some recent progress.

This script is more like a tech report rather than a thesis. Detailed introduction and ablation of some modifications are omitted as they are consistent with the papers they first appear in. We sincerely thanks them for their great contributions to this area. In the following, we'll introduce how we develop branch dev2.0 from the perspective of deployment. Some quantitative results with both accuracy and inference speed are listed to reveal the power of branch dev2.0.

\begin{figure*}[t]
		\centering
		\includegraphics[width=1.0\linewidth]{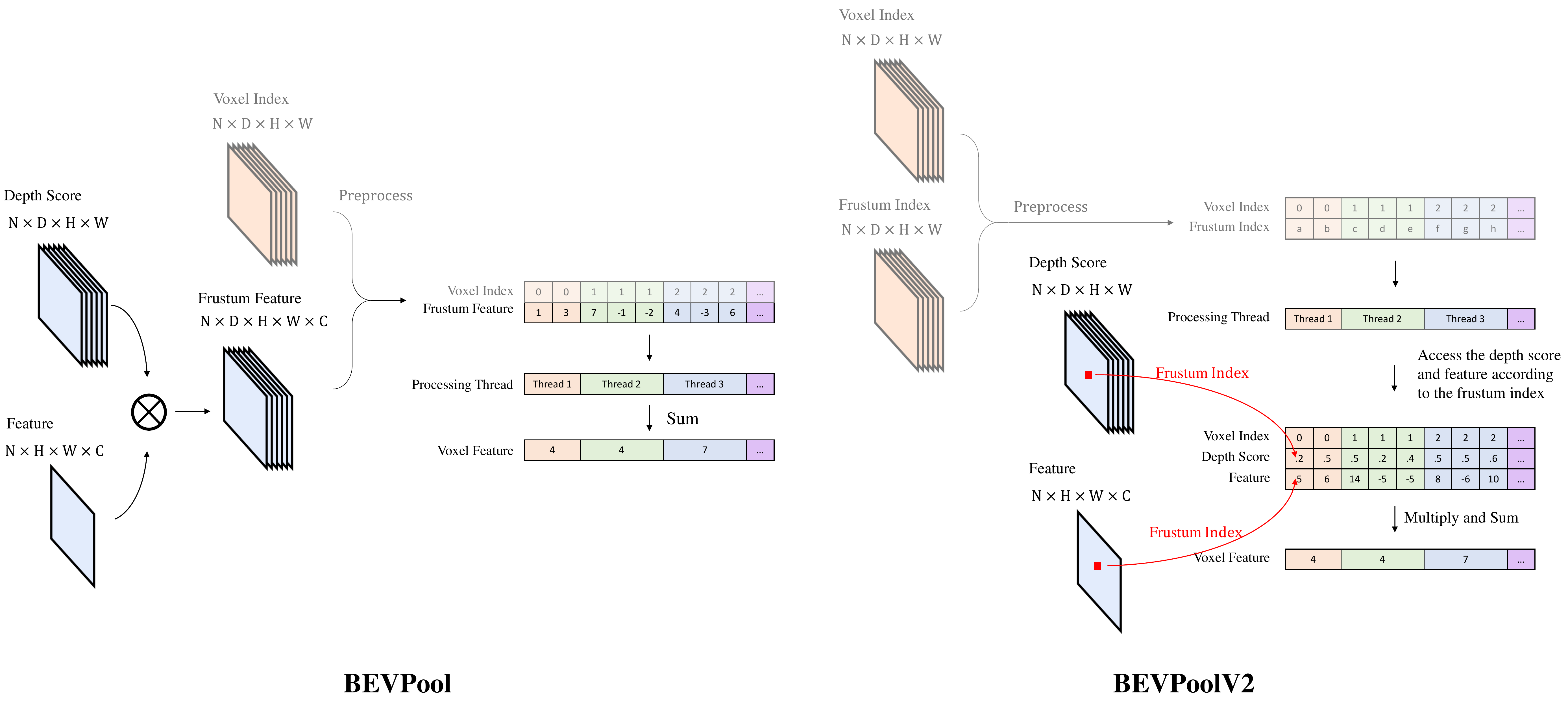}
		\caption{Illustration of BEVPool from BEVFusion \cite{BEVFusion} and BEVPoolv2 from BEVDet-dev2.0. The part in low transparency can be pre-computed offline. }
		\label{fig:viewtransformation}
\end{figure*}

\section{Modification}
\subsection{BEVPoolv2}
\begin{figure*}[t]
		\centering
		\includegraphics[width=0.85\linewidth]{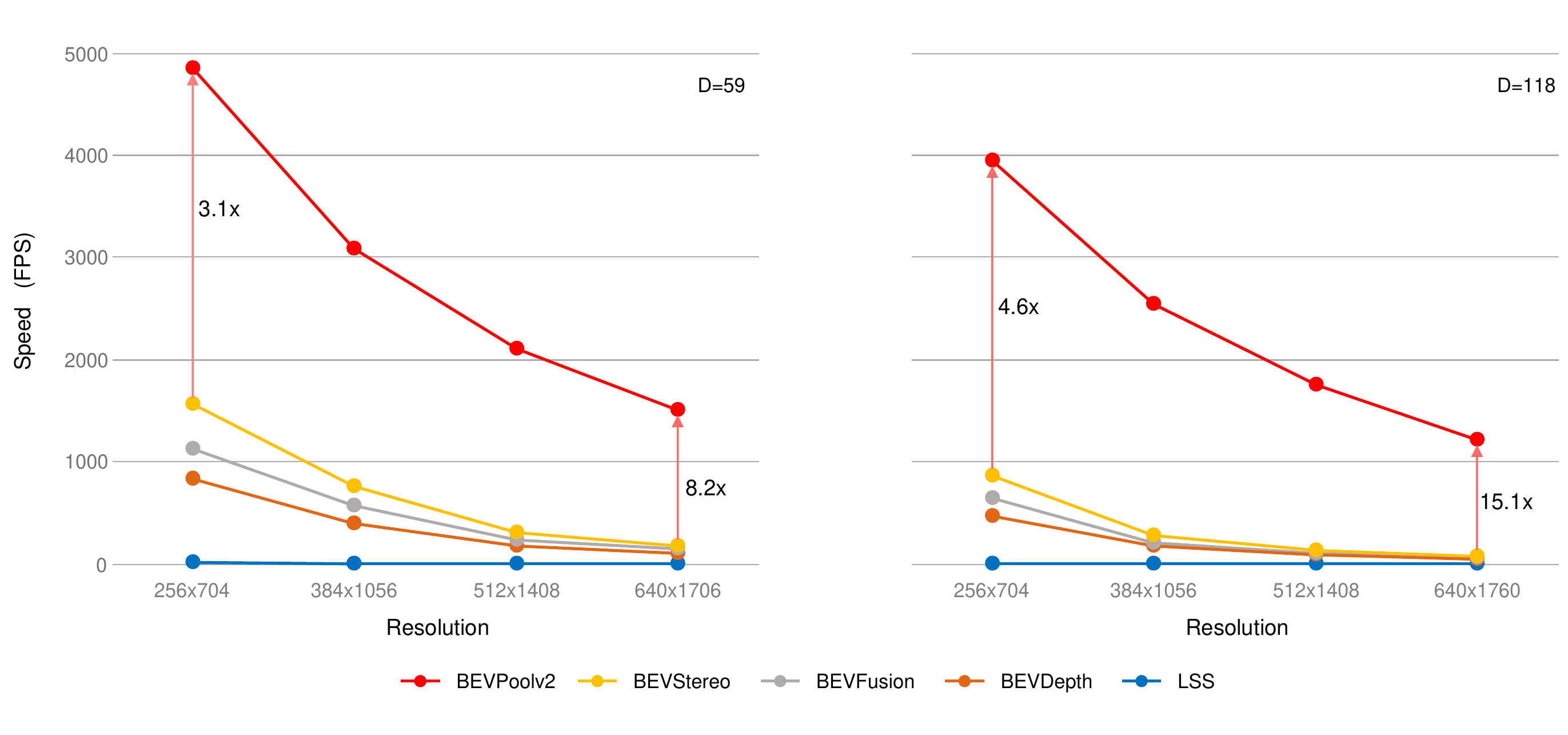}
		\caption{Inference speed of different implementation of Lift-Splat-Shoot view transformation. When the number of depth classes is set as $D=59$, the implementation of branch dev2.0, dubbed BEVPoolv2 is 3.1 times the previous fastest implementation in a low input resolution of $256\times704$ and 8.2 times in a high resolution of $640\times1760$.  }
		\label{fig:speedresolution}
\end{figure*}

\begin{figure*}[t]
		\centering
		\includegraphics[width=0.85\linewidth]{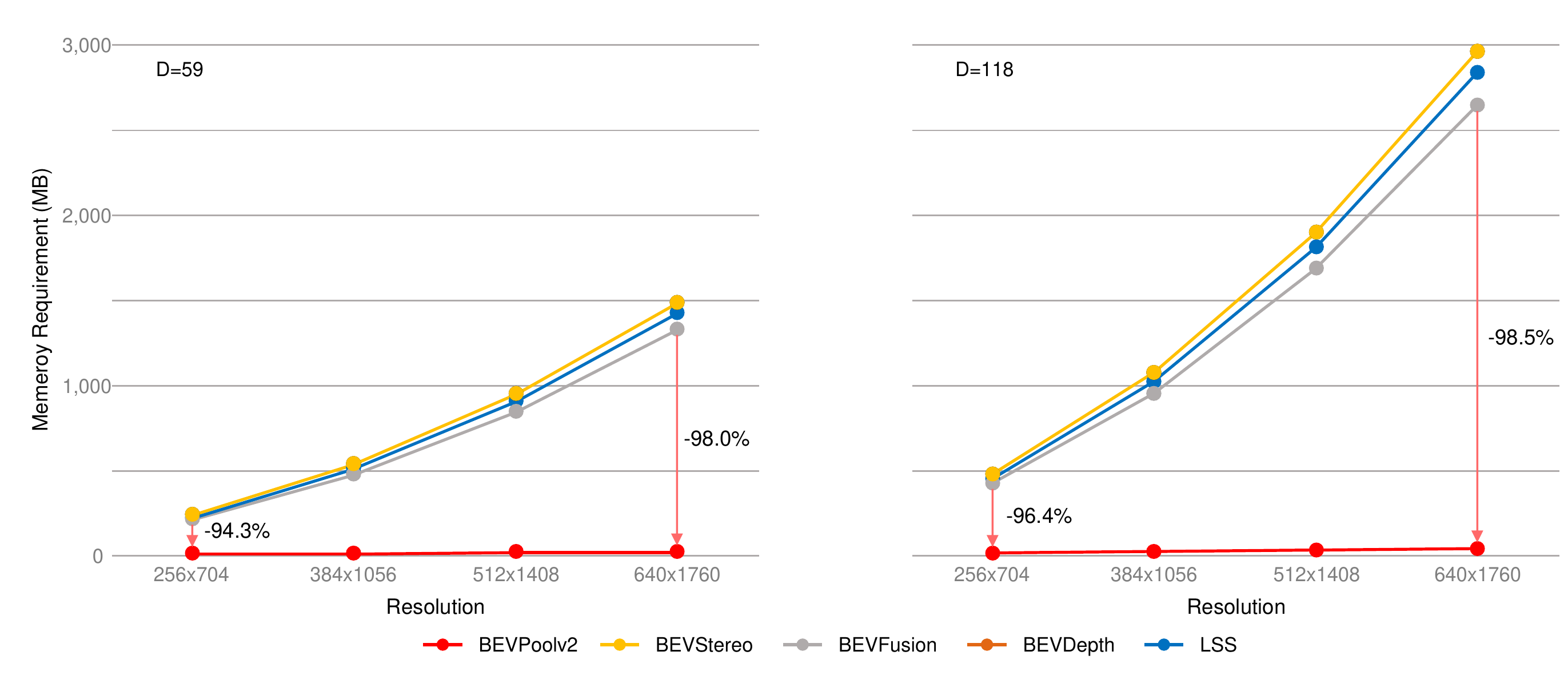}
		\caption{Memory requirement of different implementation of Lift-Splat-Shoot view transformation. When the number of depth classes is set as $D=59$, the implementation of branch dev2.0, dubbed BEVPoolv2 need just 5.7\% memory requirement of the previous fastest implementation in a low input resolution of $256\times704$ and 2.0\% in a high resolution of $640\times1760$.  }
		\label{fig:memoryresolution}
\end{figure*}

One of the main drawbacks of the Lift-Splat-Shoot view transformer \cite{LSS} is that it has to compute, store, and preprocess the frustum feature with a large size of $(N,D,H,W,C)$, where $N$, $D$, $H$, $W$ and $C$ are the number of views, the class number of depth, the height of the feature size, the width of the feature size, and the channel number of feature respectively. As illustrated in the left of Fig.~\ref{fig:viewtransformation}, it is computed with the depth score in size of $(N,D,H,W)$ and the feature in size of $(N,H,W,C)$. Then it is preprocessed alone with the voxel index, which indicates the voxel that the frustum points belong to and is computed according to the camera's intrinsic and extrinsic. The preprocessing includes filtering the frustum points that are out of the scope of all voxels and sorting the frustum points according to their voxel index. After that, the frustum points within the same voxel are aggregation by cumulative sum \cite{LSS}. Instead of cumulative sum, BEVPool in BEVFusion \cite{BEVFusion} uses multiple processing threads to accelerate this procedure. However, it still needs to compute, store and preprocess the frustum feature, which is both memory-consumptive and calculation consumptive. A similar situation can be observed in BEVDepth \cite{BEVDepth} and BEVStereo \cite{BEVStereo} with VoxelPool. As illustrated in Fig.~\ref{fig:speedresolution} and Fig.\ref{fig:memoryresolution}, when the input resolution is increased, their efficiency degenerates obviously and the memory requirement boosts to an extremely high level. For example with a depth dimension of 118 and an input resolution of $640\times1760$, the previous fastest implementation can only be processed at 81 FPS and occupies cache memory up to 2,964 MB. It is obviously infeasible on edge devices.

We propose BEVPoolv2 in branch dev2.0 with engineering optimization to completely get rid of this drawback. As illustrated in the right of Fig.~\ref{fig:viewtransformation}, the idea is very simple. We just use the index of frustum point as a trace of them to be preprocessed alone with their voxel index. Then we access the value of the frustum point feature in the processing thread according to the frustum index. In this way, we avoid explicitly computing, storing, and preprocessing the frustum feature. Thus, the memory and calculation can be saved and the inference can be further speeded up. Both the voxel index and frustum index can be pre-computed and preprocessed offline. During the inference, they just act as fixed parameters.

As illustrated in Fig.~\ref{fig:speedresolution}, with pre-computation acceleration, the inference speed of the BEVPoolv2 is up to 4,863 PFS in a low input resolution of $256\times704$ and still maintains 1,509 FPS in high resolution of $640\times1760$. It is 3.1 times the previous fastest implementation in a low input resolution of $256\times704$ and 8.2 times in a high resolution of $640\times1760$. BEVPoolv2 is so fast that makes the view transformation no longer to be the bottleneck in the hold pipeline. Last but not least, the memory saved by avoiding storing the frustum feature is also appealing. As illustrated in Fig.~\ref{fig:memoryresolution}, the memory requirement of the ours implementation is far less than that of the previous implementations and can be maintained at an extremely low value when the input solution is increasing. This makes the Lift-Splat-Shoot view transformer feasible to be deployed on edge devices.

\subsection{TensorRT}
We support the conversion of the BEVDet paradigm from PyTorch to TensorRT in our codebase. We list the inference latency of some configurations in Tab.~\ref{tab:trt} for reference. With the TensorRT-FP16 backend, BEVDet-R50 \cite{BEVDet} configuration with $256\times704$ input resolution can be processed at 138.9 FPS on an NVIDIA GeForce RTX 3090 GPU.

%\begin{table}[h]
%  \centering
%  \caption{Inference speed of different backends with BEVDet-R50 \cite{BEVDet} configuration on a NVIDIA GeForce RTX 3090 GPU. We omit the postprocessing, which runs about 14.3 ms with PyTorch backend.}
%    \resizebox{0.7\linewidth}{!}{
%    \begin{tabular}{c|ccc|cc|ccccccc|crc}
%    \hline
%
%    \hline
%    Resolution     & Backend       &Time(ms)    \\
%    \hline
%    $256\times704$ &PyTorch        & 38.6           \\
%    $256\times704$ &TensorRT       & 18.8       \\
%    $256\times704$ &TensorRT-fp16  & 7.03       \\
%    \hline
%
%    $384\times1056$ &PyTorch        & 66.7       \\
%    $384\times1056$ &TensorRT       & 27.2       \\
%    $384\times1056$ &TensorRT-fp16  & 10.2       \\
%
%    \hline
%    $512\times1408$ &PyTorch        & 105.3           \\
%    $512\times1408$ &TensorRT       & 42.0       \\
%    $512\times1408$ &TensorRT-fp16  & 14.9       \\
%    \hline
%    $640\times1600$ &PyTorch        & 138.9           \\
%    $640\times1600$ &TensorRT       & 55.2       \\
%    $640\times1600$ &TensorRT-fp16  & 19.2       \\
%    \hline
%    \end{tabular}%
%    }
%
%  \label{tab:trt}%
%\end{table}%

\begin{table}[h]
  \centering
  \caption{Inference speed of different backends with BEVDet-R50 \cite{BEVDet} configuration on an NVIDIA GeForce RTX 3090 GPU. We omit the postprocessing, which runs about 14.3 ms with the PyTorch backend.}
    \resizebox{1.0\linewidth}{!}{
    \begin{tabular}{c|ccccccccccccccc}
    \hline

    \hline
    Backend         & $256\times704$       & $384\times1056$ & $512\times1408$ & $640\times1760$   \\
    \hline
    PyTorch        & 37.9    & 64.7   & 105.7   & 154.2       \\
    TensorRT       & 18.4    & 25.9   & 40.0    & 58.3       \\
    TensorRT-FP16  & 7.2    & 10.6   & 15.3    & 21.2      \\
    \hline
    \end{tabular}%
    }

  \label{tab:trt}%
\end{table}%

\subsection{Receptive Field}
In the previous version of BEVDet \cite{BEVDet}, we inherit the receptive field from CenterPoint \cite{CenterPoint3D} without modification. The receptive field in CenterPoint takes the origin of the Lidar coordinate system as the center. However, it is inconsistent with the receptive field in evaluation, which takes the ego (IMU) coordinate system as the center. The inconsistency will lead to performance degeneration. We follow BEVDepth \cite{BEVDepth} to set the origin of the ego coordinate system as the center of the receptive field.

\subsection{Other Modifications}
\paragraph{BEVDepth} We support all techniques in BEVDepth \cite{BEVDepth} including the depth supervision from Lidar, depth correction, and camera-aware depth prediction.

\paragraph{Temporal Fusion} We introduce the long-term fusion in SOLOFusion \cite{SOLOFusion} to upgrade the temporal fusion paradigm in BEVDet4D \cite{BEVDet4D}.

\paragraph{Stereo Depth Estimation} So far, we haven't integrated any technique for stereo depth prediction like BEVStereo \cite{BEVStereo}, STS \cite{STS}, or SOLOFusion \cite{SOLOFusion}. We consider that they trade too much inference time for small accuracy improvement when compared with other factors like increasing the input resolution or increasing the backbone size. Besides, all the paradigms relied on cost volume are too complicated to be deployed to other platforms. More effort is needed in this promising direction.

\subsection{BEVDet4D-R50-Depth-CBGS}
We construct the BEVDet4D-R50-Depth-CBGS configuration as an example, which integrates all the aforementioned features. It scores 52.3 NDS and can be processed at 16.4 FPS with the PyTorch backend on an NVIDIA GeForce RTX 3090 GPU.

{\small
\bibliographystyle{ieee_fullname}
\bibliography{bevpoolv2}
}
\end{document}